\documentclass[preprint,journal]{vgtc}       

\ifpdf
  \pdfoutput=1\relax                   
  \usepackage{graphicx}                
  \DeclareGraphicsExtensions{.pdf,.png,.jpg,.jpeg} 
\else
  \usepackage{graphicx}                
  \DeclareGraphicsExtensions{.eps}     
\fi%

\graphicspath{{figures/}{pictures/}{images/}{./}} 

\usepackage{microtype}                 
\PassOptionsToPackage{warn}{textcomp}  
\usepackage{textcomp}                  
\usepackage{mathptmx}                  
\usepackage{times}                     
\usepackage{cite}                      
\usepackage{tabu}                      
\usepackage{booktabs}                  
\usepackage{color}
\usepackage{enumitem}
\usepackage[table]{xcolor}
\usepackage{makecell}
\newcommand{\revcolor}{black}

\onlineid{0}

\vgtccategory{Research}
\vgtcpapertype{system}

\newcommand{\techname}{Manifold}
\newcommand{\company}{Uber Technologies, Inc}

\title{\techname: A Model-Agnostic Framework for Interpretation \\ and Diagnosis of Machine Learning Models}

\author{Jiawei Zhang, Yang Wang, Piero Molino, Lezhi Li and David S. Ebert, \textit{Fellow, IEEE}}
\authorfooter{
\item
 Jiawei Zhang and David S. Ebert are with Purdue University, E-mail: \{zhan1486$\vert$ebertd\}@purdue.edu. This work was done while the first author was at Uber Technologies, Inc.
\item
 Yang Wang and Lezhi Li are with Uber Technologies, Inc, E-mail: \{gnavvy$\vert$lezhi.li\}@uber.com.
 \item
 Piero Molino is with Uber AI Labs, E-mail: piero@uber.com.
}

\abstract{
Interpretation and diagnosis of machine learning models have gained renewed interest in recent years with breakthroughs in new approaches. We present \techname, a framework that utilizes visual analysis techniques to support interpretation, debugging, and comparison of machine learning models in a more transparent and interactive manner. Conventional techniques usually focus on visualizing the internal logic of a specific model type (i.e., deep neural networks), lacking the ability to extend to a more complex scenario where different model types are integrated. To this end, \techname\ is designed as a generic framework that does not rely on or access the internal logic of the model and solely observes the input (i.e., instances or features) and the output (i.e., the predicted result and probability distribution). We describe the workflow of \techname\ as an iterative process consisting of three major phases that are commonly involved in the model development and diagnosis process: inspection (hypothesis), explanation (reasoning), and refinement (verification). The visual components supporting these tasks include a scatterplot-based visual summary that overviews the models' outcome and a customizable tabular view that reveals feature discrimination. We demonstrate current applications of the framework on the classification and regression tasks and discuss other potential machine learning use scenarios where \techname\ can be applied.
} 

\keywords{Interactive machine learning, performance analysis, model comparison, model debugging}

\CCScatlist{ 
 \CCScat{K.6.1}{Management of Computing and Information Systems}%
{Project and People Management}{Life Cycle};
 \CCScat{K.7.m}{The Computing Profession}{Miscellaneous}{Ethics}
}

\teaser{
  \centering
  \includegraphics[width=.95\linewidth]{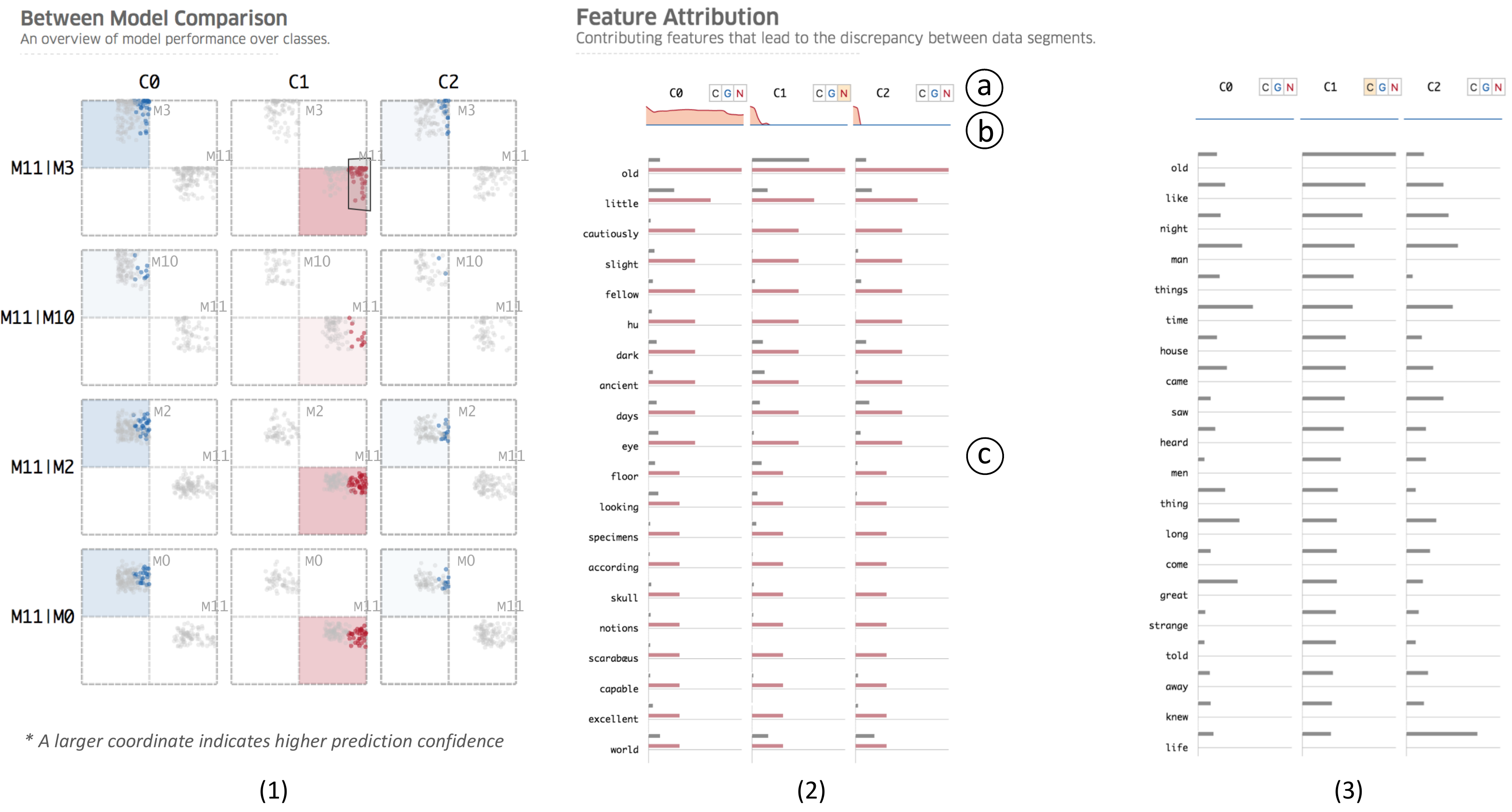}
  \caption{Manifold consists of two interactive dialogs: a model comparison overview (1) that provides a visual comparison between model pairs using a small multiple design, and a local feature interpreter view (2) that reveals a feature-wise comparison between user-defined subsets (c) and provides a similarity measure (b) of feature distributions. The user can sort based on multiple metrics (a) to identify the most discriminative features among different subsets, i.e., sort based on the selected subset in red (2) or a specific class such as $C_1$ (3).}
	\label{fig:interpreter}
}

\vgtcinsertpkg

\begin{document}



\firstsection{Introduction}
\maketitle
Recent technical breakthroughs in the machine learning field have led to highly improved accuracies and utilization in many scenarios, including sophisticated pattern recognition tasks~\cite{lecun2015deep, abadi2016tensorflow}. However, these technical advances pose two major challenges. First, the complexity of the models being designed and adopted has significantly increased to the point that is difficult for model developers to explain why and how the model works. Second, model developers often lack solid reasoning or evidence to guide their development and debugging due to the hidden mechanisms of the models, making this iterative process more time-consuming and error-prone. Both of these challenges require more effective approaches that enable interpretation and explanation of machine learning processes~\cite{ribeiro2016should, kulesza2015principles, Krause2016interacting}.

Visual and interactive interfaces have proved to be effective in terms of enabling users to integrate domain knowledge in the process of interpreting and diagnosing these complex models~\cite{fekete2013visual, tam2017analysis, muhlbacher2014opening, amershi2014power}. Typical solutions include visualizing the internal structure or intermediate states of the model to enhance the understanding and interpretation~\cite{wongsuphasawat2018visualizing, ming2017understanding, kahng2018cti}, evaluating and analyzing the performance of models or algorithms~\cite{ren2017squares, amershi2015modeltracker, liu2018visual}, and interactively improving the models at different development stages such as feature engineering or hyperparameter tuning through integration of domain knowledge~\cite{paiva2015approach, zhao2014lovis, brooks2015featureinsight}. Nevertheless, the focus of these approaches has been mostly restricted to a specific model type or task type (i.e., classification tasks), lacking the ability to extend to more complex industry-level use scenarios where the size and the complexity of both the model and the task increase.

In this paper, we present an interactive framework called \techname\ to address these problems of integrating, evaluating and debugging multiple machine learning models. The design process of the framework has been guided by three major phases that are typically involved in diagnosing and comparing machine learning models: inspection (hypothesis), explanation (reasoning), and refinement (verification). The \techname\ interface supports these phases through two main visual components. First, we design a novel scatterplot-based visual technique that provides a comparative visual summary of the diversity and complementarity of the model pairs, and allows the users to effectively inspect \textit{symptom} data instances and make hypotheses accordingly. The technique consists of multiple encoding schemes that are flexible and adaptable to various task types such as classification or regression. Second, we design a tabular view for the users to visually discriminate features extracted from \textit{symptom} instances and identify which features are more influential in the models' outcome, thus providing explanations for the hypotheses generated earlier on. These explanations can then be incorporated into a new iteration of the model development in order to validate and refine the model.

\textcolor{\revcolor}{Comparing to state-of-the-art solutions in this area, we focus on generality as the primary property of the framework. \techname\ is model-agnostic, in the sense that it does not need access to the internal logic of the model and only relies on the input instances and the output results, allowing the framework to support a broad range of model types, as long as they target the same machine learning task and have a consistent format of input and output. Furthermore, \techname\ is built upon scalable WebGL-based rendering frameworks~\cite{wang2017deckgl, lumagl} and consists of several visual summarization and interaction designs, making it possible to handle large-scale input instances while reducing potential computational and cognitive overload. In this paper, we describe the usage of the framework on two typical supervised learning tasks, multi-class text classification and regression, and discuss ideas for extending \techname\ to a broader range of machine learning tasks.}

\begin{table*}[t]
  \small\sf
  \centering
  \vspace{2mm}
  \begin{tabular}{l|p{14.5cm}}
    \toprule
    High-level phases & Low-level design goals \\
    \midrule
    Inspection (Hypothesis)   &
    \vspace{-3mm}
    \begin{itemize}[noitemsep,topsep=0pt]
    \item[\textbf{T1.1}] Provide an overall summary of results generated by multiple models.
    \item[\textbf{T1.2}] Enable an effective comparative analysis on model pairs ($M_i$ vs $M_j$):
    \begin{itemize}[noitemsep,topsep=0pt]
    \item[\textbf{T1.2.1}] Which model between $M_i$, $M_j$ has a higher accuracy among all test instances?
    \item[\textbf{T1.2.2}] On which instances does $M_i$ (or $M_j$) make a correct prediction but  $M_j$ (or $M_i$) fail?
    \item[\textbf{T1.2.3}] On which instances does $M_i$ and $M_j$ make an agreement (both correct or both incorrect)? If both of them are correct, which model generates higher prediction scores (more confident)?
    \end{itemize}
    \item[\textbf{T1.3}] Enable an effective comparative analysis on a model and others ($M_i$ vs the rest):
    \begin{itemize}[noitemsep,topsep=0pt]
    \item[\textbf{T1.3.1}] On which instances does $M_i$ make an agreement (disagreement) with the rest of the models?
    \item[\textbf{T1.3.2}] Which models overall have a similar (different) behavior as $M_i$ in terms of the model outcome?
    \end{itemize}\vspace*{-\baselineskip}
    \end{itemize}
    \\ \hline
    Explanation (Reasoning)   &
    \vspace{-2mm}
    \begin{itemize}[noitemsep,topsep=0pt]
    \item[\textbf{T2.1}] Provide a visual summary for the feature distribution of the user-defined instance subset.
	\item[\textbf{T2.2}] Enable an effective visual comparison of two different instance subsets regarding the feature distribution:
     \begin{itemize}[noitemsep,topsep=0pt]
     \item[\textbf{T2.2.1}] Allow comparison of individual features.
     \item[\textbf{T2.2.2}] Enable a quick Identification of the most discriminative features.
     \item[\textbf{T2.2.3}] Provide a quantitative measure of the overall distribution similarity.
     \end{itemize}\vspace*{-\baselineskip}
     \end{itemize}
     \\
     \hline
    Refinement (Verification) &
    \vspace{-2mm}
    \textbf{T3} \hspace{.2mm} Generate feature (or model architecture) encoding strategies and iterate model refinement.
	\\
    \bottomrule
  \end{tabular}
  \vspace{1mm}
  \caption{A high-level domain characterization including inspection (hypothesis), explanation (reasoning), and refinement (verification), and corresponding low-level task designs for the \techname\ framework.}
  \label{tab:tasks}
  \vspace{-4mm}
\end{table*}

\section{Related Work}

Most research on visually interpreting deep learning models requires access to the internal working mechanism of the model itself, for example, visualizing and understanding the intermediate calculation or the internal structures of the model~\cite{wongsuphasawat2018visualizing, ming2017understanding, kahng2018cti, yosinski2015understanding, liu2017towards, strobelt2018lstmvis, rauber2017visualizing, liu2018analyzing}. Since our work aims to support the diagnosis of a broader range of models without relying on their internal logics, in this section we do not discuss in detail the work within this category. Instead, we review these more relevant directions: model diagnosis and debugging, model performance analysis and interactive model refinement.

\subsection{Model Debugging and Performance Analysis}
Summary statistics (i.e., accuracy, F-measure, confusion matrices) for performance analysis are usually representative of a coarse-grained perspective of the performance especially when evaluating multiple models, and can potentially lead to a biased cognition. Research has explored providing an exploratory environment of model performance at a fine-grained level. Alsallakh et al.~\cite{alsallakh2014visual} propose \textit{Confusion Wheel} that arranges different classes based on a radial layout and use histograms to show the statistics of the true/false positive/negative associated with each class and their prediction confidence. However, the technique lacks the ability to support effective comparison of multiple models. Amershi et al.~\cite{amershi2015modeltracker} propose \textit{ModelTracker} that directly encodes the model's prediction score for each instance using its Cartesian coordinates in the 2D space such that the instances of similar prediction scores have spatial proximity. Similarly, the efficiency of this technique remains unknown when applied to multiple models or multi-class classification tasks. To solve these challenges, Ren et al.~\cite{ren2017squares} present \textit{Squares}, a performance analysis system that juxtaposes a set of histograms to present the prediction scores in a multi-class classification task and allows the user to investigate different models by visually comparing multiple histograms.
\textcolor{\revcolor}{However, since the results generated by multiple models are presented separately in the visual space and there does not exist a visual indication of how different the models behave on the same instance, the technique lacks the ability to characterize the model diversity at the instance level. In contrast, our technique supports more fine-grained functionality to drill down to specific \textit{symptom} instances on which different models agree or disagree. Furthermore, our technique can be applied to both classification and regression tasks.}

Other research directions aim to identify important features and instances that are relevant to an issue within the model~\cite{Cadamuro2016Debugging, Krause2016interacting, krause2017workflow, kulesza2015principles, kulesza2010explanatory}. Cadamuro et al.~\cite{Cadamuro2016Debugging} present a conceptual analysis and diagnosis loop that allows end users to iteratively detect bugs, identify root cause (the training instances that contribute to the bug the most) and resolve the issue. Liu et al.\cite{liu2018visual} suggest a diagnosis system for the training process of tree boosting methods. Krause et al.~\cite{Krause2016interacting} design a system called \textit{Prospector} for understanding how a specific feature contributes to the prediction by adjusting the feature value and examine the corresponding change of the predicted result. Krause et al.~\cite{krause2017workflow} propose a model diagnosis workflow that identifies a set of features that tend to influence the model outcome on a single instance the most. The instances that have the same set of influential features are then aggregated and summarized in a tabular display for effective inspection. \textcolor{\revcolor}{Since our work primarily focuses on the comparison of multiple models, our approach slices data and creates visual summaries based on model-level properties (i.e., model agreement or disagreement) instead of the feature-level properties (i.e., influential or sensitive features) that are commonly adopted in the aforementioned techniques. Once the user selects specific data slices (\textit{symptom} instances), our approach provides summary statistics and visual comparison at the feature level, allowing them to identify the most discriminative features and generate explanations accordingly.}

\subsection{Interactive Model Refinement and Ensemble}
Human-in-the-loop approaches facilitate the integration of the end users' knowledge in the process of supervising and refining models. The knowledge being integrated in the model is usually acquired by the end users either through prior experience or by interactively examining model outcome and intermediate states during the analysis process. Typical solutions include improving hyperparameters~\cite{kapoor2010interactive, choo2013utopian}, features~\cite{brooks2015featureinsight, zhao2014lovis, muhlbacher2013partition}, or training instances~\cite{paiva2015approach} and investigating multiple models in order to acquire an optimal model ensemble~\cite{talbot2009ensemblematrix, schneider2017visual, zhao2014lovis}.

Paiva et al.~\cite{paiva2015approach} suggest an incremental classification scheme that allows the users to observe training instances using a similar-based layout method and specify which samples to use for the learning process, hence improving the overall quality of the training set. Brooks et al.~\cite{brooks2015featureinsight} propose \textit{FeatureInsight} for improving the feature engineering through interactive visual summaries. The system allows a feature-level comparison between the wrongly predicted instances (errors) and the correctly predicted instances (contrasts), recommending features that could potentially be used to reduce erroneous predictions. While \textit{FeatureInsight} is mainly applied to binary classification tasks, our system can be applied to a large range of use cases including multi-class classification and regression. Talbot et al.~\cite{talbot2009ensemblematrix} propose \textit{EnsembleMatrix} that juxtaposes the confusion matrix of multiple classifiers for visual comparison and supports linear combinations of these classifiers for more effective model ensemble. Zhao et al.~\cite{zhao2014lovis} propose \textit{LoVis} that partitions instances into segmentations and analyzes which models or features produce better predictions for each local data segment at multiple levels of granularity. Schneider et al.~\cite{schneider2017visual} juxtapose the model space and the data space in the visual interface and allows the end user to filter on these two orthogonal spaces. Compared to this approach, our system combines the model and data space in the same visualization through a compact small multiple based visual design, thus reducing the interaction overload and easing the selection of erroneous instances.

\section{Domain Characterization}
The iterative design process of \techname\ has been a collaborative effort between machine learning researchers and engineers and visualization researchers from a ride-sharing company. In this section, we first discuss the background of this work based on iterative conversations with the domain experts. Then we characterize abstractions of the tasks that the framework should support using visual analysis vocabularies~\cite{munzner2009nested}.

\subsection{Motivation}
The challenges that machine learning researchers and developers face when developing new machine learning models are characterized into the following aspects.

\textbf{Debugging coding errors in the model:} Many model failures can be caused by coding errors that lie in several different aspects: errors in the code implementation, errors in the mathematical foundation the implementation is based upon, and errors in the data preprocessing stage. Although relatively difficult to figure out in the code, those errors usually have catastrophic implications for the model performance. Hence, it is relatively easy to identify them when the results and performance measures are presented to the user.

\textbf{Understanding strengths and weaknesses of one model both in isolation and in comparison with other models}: This is usually carried out by identifying under which scenario a model  performs poorly (i.e., returns erroneous prediction results or results that are far from the targeted value, or returns results that are inconsistent with other models). Inspecting those scenarios can be difficult especially when the size and complexity of the instance and feature space become significant (i.e., instances that contain a large amount of text, a large number of numerical features, or a combination of several different types of features). Thus, the domain experts require techniques that can not only summarize relevant information at a high level, but also allow drilling down to a subset of data points based on user-defined conditions.

\textbf{Model comparison and ensembling}: Effective comparative analysis of multiple models not only helps assess which model works better in specific scenarios, but also provides insights into model ensembling, a practical strategy that combines complementarity models in order to achieve higher accuracy than any individual one. Typical statistical measures for performance analysis provide an overall understanding of the quality of the prediction. However, they do not characterize the differences in terms of the types of error the models make (i.e., in the case of a multi-class classification task, on which class does a model work better than another and why?). Gathering these insights becomes more challenging as the number of models to compare increases, in which case a visual analysis environment can be extremely helpful since end users can interactively supervise each step of the analysis process in order to reduce the complexity.

\textbf{Incorporating insights gathered through inspection and performance analysis into model iterations}: Although the limitations of the model have been identified, addressing these issues may not be straight-forward and requires a comprehensive understanding and creativity in terms of both the model's working logic and the domain knowledge on the input dataset. Visual analysis techniques can be helpful in both aspects. End users can apply appropriate visualizations to decouple complex structure within the model and enhance the understanding. An exploratory analysis environment can help users explore the large-scale and complex input data in order to gain additional insight into the data and hence uncover relationships between the data and model output.


\subsection{Task Characterization and Design Goals}
We characterize three high-level analysis phases that are commonly involved in the model diagnosis and comparison process. We note that in practical scenarios, these phases are not easily separable and can occur concurrently at some point during the analysis process. However, we summarize them individually for the sake of better characterization and illustration.
We then detail the low-level task design goals which \techname\ has to fulfill in each phase~\cite{munzner2009nested} in Table~\ref{tab:tasks}.

\textbf{Inspection (Hypothesis)} is the entry of the analysis process when the user designs a model and attempts to investigate and compare the model outcome with other existing ones. During this phase, the user compares typical performance metrics, such as accuracy, precision/recall, and receiver operating characteristic curve (ROC), to have coarse-grained information of whether the new model outperforms the existing ones. Then the user focuses on a specific instance subset of interest in order to narrow down her analysis space~\cite{krause2017workflow, kahng2018cti}. Typical approaches to sampling an instance subset include:
	\begin{itemize}[itemsep=0em,topsep=0pt]
  \item The user has reasonable knowledge of the instances (i.e., feature distribution, ground truth) prior to the analysis process. This type of subset makes it easier for the user to make sense of the model outcome and correlate it with the input (described in the explanation phase).
  \item The user filters a subset of instances that have some features in common, which is typically done through a faceted search on the input features or the meta data information. Similarly, this approach samples a subset where feature distributions are in a simpler form and explicit to the user,  thus easing the interpretation of the instances.
  \item The user identifies a subset where the results generated by the model are erroneous or suspicious, for example, the instances where the new model has low accuracy while others have high accuracy. We define this type of subset as a \textit{symptom} set since it is representative of a potential fault within the model. The \textit{symptom} set is of particular interest to the user during the diagnosis process.
  \end{itemize}
Once the user finalizes the selection, she can make hypotheses of potential issues within the model that may lead to the erroneous result and proceed to the explanation phase.

\textbf{Explanation (Reasoning):} After selecting a \textit{symptom} set, in this phase, the user attempts to explain her hypotheses. The user may access the detailed information at the instance level or the feature level that can potentially explain the \textit{symptom}. Comparative analysis is intensively involved in this phase. For example, after selecting a subset of false positive instances (the ground truth class is A while the result predicted by the model is B), the user may want to investigate what features or local structures of the selected subset are more similar to the set which ground truth class is B and less similar to the set which ground truth class is A. These features could be influential to the false positive results and hence are regarded as an explanation of the \textit{symptom}. We note that an explanation may not necessarily have a causal relationship with the \textit{symptom}. The user can generate multiple explanations relevant to a \textit{symptom}. A verification phase is required to validate the explanation.

\textbf{Refinement (Verification):} In this phase, the user attempts to verify the explanations generated from the previous phase through encoding the knowledge extracted from the explanation into the model and testing the performance. This verification process may be both expensive and challenging. First, depending on the model type, the user can either apply feature engineering strategies or adjust the internal architecture of the model (typically for deep neural networks). Hence, the user needs to have a comprehensive understanding of the model's internal mechanism. Otherwise, it becomes hard for her to derive a reasonable encoding strategy based on the explanation. Second, since another training and testing session is required to test the model, the phase can be tedious and time-consuming.

\begin{figure*}[t]
 \centering 
 \includegraphics[width=.9\textwidth]{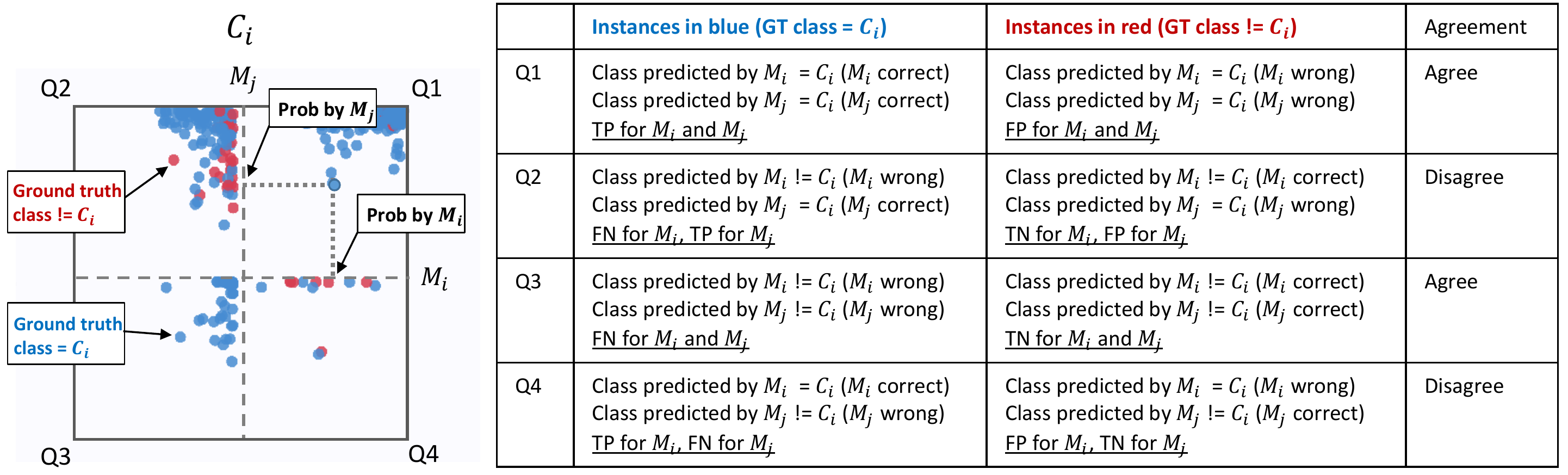}
 \caption{Left: The visual encoding strategy of the prediction results for the model pair $M_i$ and $M_j$ on the class $C_i$. Right: the interpretation of instances in each quadrant in the matrix cell (TP: true positive; TN: true negative; FP: false positive; FN: false negative). The instances in the first (Q1) and third (Q3) quadrants indicate that $M_i$ and $M_j$ agree on the predictions while those in the second (Q2) and fourth (Q4) quadrants indicate the two models disagree on the predictions. }
 \label{fig:encoding}
\end{figure*}

\begin{figure*}[tb]
 \centering 
 \includegraphics[width=.9\textwidth]{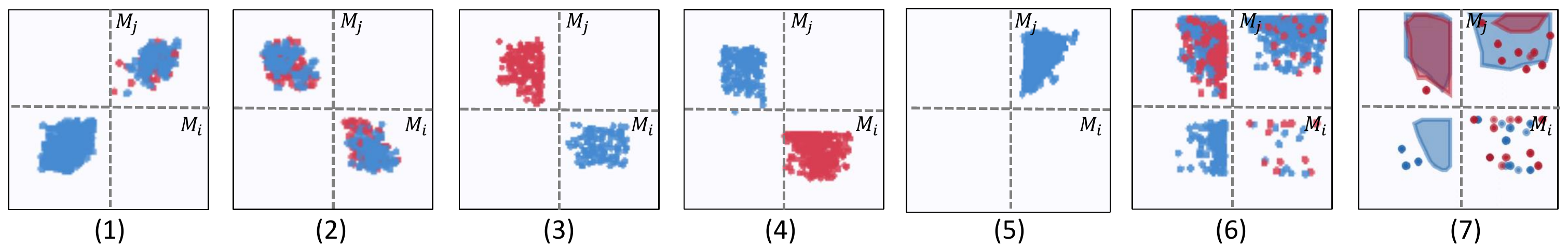}
 \caption{(1): Instances in Q1 and Q3 indicate $M_i$ and $M_j$ produce consistent predictions (The two models agree). (2) Instances in Q2 and Q4 indicate $M_i$ and $M_j$ produce inconsistent predictions (The two models disagree). (3): Instances on $M_i$ is correct while $M_j$ is incorrect. (4): Instances on which $M_i$ is incorrect while $M_j$ is correct. (5): Although the two models agree and both of them are correct, $M_j$ has a higher overall prediction probability than $M_i$, indicating $M_j$ is more confident about the prediction. (6) and (7): Comparison between the scatterplot-based visualization (6) and the contour-based visualization (7). The contour visualization reduces the overplotting issue when rendering a large set of points in the limited space (i.e., a cluster of red points in the right top corner in Q1 is clearly revealed in the contour view, however, was not shown in the scatterplot view).}
 \label{fig:example-cell}
\end{figure*}

\section{The \techname\ Framework}
The interface of \techname\ mainly consists of two visual dialogs: a model comparison overview shown in Figure~\ref{fig:interpreter}(1) for visually inspecting potential issues within the model and identifying \textit{symptom} instances, and a feature interpretation view shown in Figure~\ref{fig:interpreter}(2) for comparing feature distributions and generating explanations for the issue. In this section, we discuss how these two components are designed and coordinated in order to enable iterative diagnostic and comparative analysis.

\subsection{Model Comparison Overview}
In a particular machine learning task, the prediction results generated by multiple models on a set of instances usually form a high-dimensional information space that contains heterogeneous data types (boolean, categorical, numerical). For example, in the context of multi-class classification, each model predicts a class (categorical) for an instance. The prediction is associated with a probability score (numerical) that describes the confidence of the decision. The predicted class can either be consistent with the ground truth (GT) class of the instance or not, indicating the prediction is correct or incorrect (boolean). \techname\ visualizes this multi-dimensional space with an emphasis on the visual comparison of \textbf{model pairs} as pair-wise comparison is an intuitive form of comparison and requires relatively little cognitive workload. The knowledge gathered from multiple model pairs can then be composed to generate a holistic understanding of the entire model space.

Inspired by the table-based approaches for multi-dimensional data exploration~\cite{stolte2002polaris}, we propose a small multiple visualization that supports an effective visual comparison of model pairs. By encoding two orthogonal information dimensions using the row and the column of the small multiple matrix, the entire information space is naturally decomposed into subsets that are arranged in each matrix cell. In \techname, we use the row to encode the model pair and use the column to encode an orthogonal dimension depending on the use scenario and the user's preference. Without loss of generality, we describe the encoding schemes corresponding to two common supervised learning tasks: multi-class classification and regression as follows.

\subsubsection{Multi-Class Classification}
In multi-class classification, the end user is usually interested in diagnosing the model's performance on different classes. It is intuitive to represent each class by a column in the small multiple matrix. As Figure \ref{fig:encoding} shows, a single matrix cell encodes the prediction results relevant to the model pair ($M_i, M_j$) and the class $C_i$. Since we focus on not only the correctness of each model (T1.1), but also whether the two models agree or disagree on specific instances and how confident they are about the prediction (T1.2, T1.3), we adopt a Cartesian coordinate system within the 2D space of the cell in which the $X$ axis represents $M_i$ and the $Y$ axis represents $M_j$. Each point in the coordinate system represents one input instance and the coordinate on the $X$ ($Y$) axis indicates the prediction score generated by the model $M_i$ ($M_j$) on the class $C_i$. Hence, the points that are close to the origin indicate lower prediction confidence than those far from the origin. Since the prediction score is non-negative, we use the positive half and the negative half of the coordinate system to encode whether the prediction result on the instance is $C_i$ or not. For example, instances in the positive half of the $X$ axis (Q1 and Q4 in Figure \ref{fig:encoding}) indicates that they are predicted by $M_i$ as $C_i$. In contrast, instances in the negative half of the $X$ axis (Q2 and Q3 in Figure \ref{fig:encoding}) indicates that they are predicted by $M_i$ as another class rather than $C_i$. Moreover, instances in the fourth quadrant (Q4 in Figure \ref{fig:encoding}) indicates that they are predicted by $M_i$ as $C_i$, but predicted by $M_j$ as another class instead (since Q4 is within the negative half of the Y axis). The instances are color-coded according to their ground truth (GT) class. If the GT class of the instance is $C_i$, it is rendered in blue. Otherwise, it is rendered in red. Other color schemes can be applied as well, for example, using a qualitative color scheme to distinguish between different classes~\cite{ren2017squares}. However, this could generate cognitive overload when the number of classes increases. Hence, we use a red-blue color scheme by default.

As Figure \ref{fig:encoding}(right) shows, adopting the Cartesian system spanned by two models essentially slices the instances according to the model correctness including true positive (TP), true negative (TN), false positive (FP), and false negative (FN). Meanwhile, this encoding scheme explicitly expresses on which instances the two models agree (Q1 and Q3) or disagree (Q2 and Q4) and how confident they are about the prediction. Figure \ref{fig:example-cell} provides several examples for interpreting the visual encodings for the matrix cell. The interface of \techname\ provides a filter panel for the user to narrow down to an instance subset of interest, for example, the instances of which the GT class is (blue) or is not (red) $C_i$; The instances that are correctly predicted by one model but wrongly predicted by the other. In addition, when the number of model pairs, classes or instances involved in the classification task become large, the user can filter a subset of them in order to reduce the size of the small multiple matrix. The filtering operations are implemented in the following three dimensions: column-wise (class), row-wise (model pair), and cell-wise (instance). The user can filter to only view several classes among which the between-class confusion is relatively high (column-wise). The use can also filter to show the comparison between a specific model $M_i$ and other models (row-wise), in which case the number of model pairs to display is reduced from a square scale to a linear scale. The user can also choose to show a subset of instances in each cell (cell-wise). Based on our conversation with the domain experts, we found the following three filtering options are particularly useful when the user examines the cell corresponding to the model pair ($M_i$ and $M_j$) and the class $C_i$. Therefore, these three modes are displayed in the control panel by default.

\begin{itemize}[leftmargin=.4in,itemsep=0em,topsep=0pt]
  \item[ALL] The entire input instances including TP, TN, FP, and FN.
  \item[UNION] The instances which satisfy at least one of the three conditions: (1) the GT class is $C_i$; (2) the class predicted by $M_i$ is $C_i$; (3) the class predicted by $M_j$ is $C_i$. Compared to the ALL mode, the TN instances for both models are excluded in this mode.
  \item[GT] The instances of which the GT class is $C_i$ (including TP and FN instances).
\end{itemize}

This comparison overview supports two point selection methods: a quadrant selection that allows selecting all instances within a quadrant, and a lasso selection that allows selecting instances within a user-defined polygon. When the user selects a set of instances in a matrix cell, the same instances in other cells are highlighted accordingly. Moreover, the background of the quadrants that contain the highlighted instances is rendered based on a linear interpolation between red and blue to indicate the portion of red and blue instances within the selection, as shown in Figure~\ref{fig:interpreter}(1). When the number of points rendered in the matrix cell becomes larger, the scatterplot visualization can potentially suffer an overplotting issue and generate visual confusion. \textcolor{\revcolor}{As a design alternative, we integrate a contour visualization that provides a visual abstraction of the scatterplot in order to support scalability in data size and complexity.} The contours are generated by detecting dense clusters of the points in each quadrant and then computing the concave hulls for each cluster. As Figure \ref{fig:example-cell}(6,7) shows, the contour visualization clearly reveals the distribution of the instances in red and blue, which is difficult to read in the scatterplot since the points overlap each other.

\subsubsection{Regression}
\label{des:regression}
In a typical regression task, the model usually outputs a continuous variable instead of a discrete label as in the classification task. Hence, residual analysis is typically used to evaluate model performance, which is defined as the difference $\epsilon$ between the predicted value $\hat y$ and the observed value $y$ (GT value) of the output variable ($\epsilon = \hat y$ - y). As shown in Figure~\ref{fig:regression-result}, for each instance we encode this residual value to its coordinate in the Cartesian system. The positive and negative half of the axis naturally depicts whether the predicted value is higher (over-predict) or lower (under-predict) than the GT value. Hence, in this encoding scheme the points that are close to the origin indicate a lower error compared to those that are far from the origin. We note that this interpretation is different from the case of the classification task, where the points near the origin indicate a lower prediction score.

As regression tasks usually do not have discrete output labels, the user can customize the data slicing based on other dimensions he is interested in, such as a specific metadata attribute associated with the input instances, and encode that dimension using the column of the matrix. This is especially helpful when diagnosing a complex regression task that is composed of several subtasks of smaller scales. For example, the estimated time of arrival (ETA) of a food delivery service can consist of individual components include food preparation, food transportation and customer pick up. Decomposing the ETA prediction tasks into these subtasks and investigating them individually can help reduce the complexity of the overall task and enable an inspection of performance and comparison at a fine-grained level.

\subsection{Feature Interpretation View}
We describe the feature interpretation view within the context of multi-class text classification. We assume that trigrams (n-grams when n equals 3)~\cite{manning1999foundations} are used as text features and several common linguistic analysis methods such as term frequency (TF) or term frequency-inverse document frequency (TF-IDF) are used to generate the feature value. This component also accommodates other feature types, i.e., numerical and categorical features in Figure~\ref{fig:regression-result}(right).

Once the user selects a \textit{symptom} set in the model comparison overview, for example, the selected instances in red in Figure~\ref{fig:interpreter}(1), the feature interpretation view provides a visual comparison of feature distributions between the selected instances and instances belonging to each class. As Figure \ref{fig:interpreter}(2) shows, this view presents a tabular view where each row represents a feature (a trigram in this case) and each column represents a class, as is consistent with the model comparison overview (T2.2.1). The length of the red (blue) bar in the grid cell encodes the aggregated value of the features from the red (blue) instances within the selection,  revealing the prominence of features of selected instances.
The length of the gray bar encodes the aggregated value of the features from the instances belonging to the class of the corresponding column, representing the feature distribution of the individual class.

The area of the line chart on top of each column (Figure~\ref{fig:interpreter}b) encodes the Kullback-Leibler divergence (KL-divergence)~\cite{kullback1951information} of the two distributions within the column, indicating which class has a similar feature distribution to the selected instances (T2.2.3). For example, Figure~\ref{fig:interpreter}c clearly shows that the distribution of the red bars is more similar to $C_0$ (the first column) than $C_1$ (the second column) and $C_2$ (the third column), with the KL-divergence chart showing a consistent result. The user can sort in descending order and show top K features (T2.2.2) using a button group (Figure \ref{fig:interpreter}a). Clicking on one button sorts based on the corresponding bar in the column (C: gray bar -- class related; G: blue bar -- GT instances; N: red bar -- non-GT instances). This sorting operation allows the user to identify prominent features of the selected subset. For example, in Figure \ref{fig:interpreter}(3), sorting features based on Class $C_1$ clearly shows highly frequent keywords such as \textit{old}, \textit{like}, and \textit{night}. A visual comparison in the same row indicates they are less frequent in the other two classes. Clicking on two buttons sorts based on the difference of the two bars, allowing the user to identify which features are the most discriminative features between the two subsets.

\section{Case Study}
\textcolor{\revcolor}{We present two case studies to showcase how \techname\ can facilitate reasoning in different usage scenarios. Due to data sensitivity, in both studies we used publicly available datasets instead of the company-specific datasets for illustration. To further demonstrate the efficacy of the system, we interviewed the machine learning researchers and engineers who were involved in the project and used the system for company-specific use cases and present their feedback.}

\subsection{Multi-Class Classification}
We use the spooky author identification dataset from Kaggle~\cite{spookyauthor} to showcase the use of \techname\ in the multi-class classification. The dataset contains a set of excerpts from horror stories written by three authors (Edgar Allan Poe (EAP), Mary Shelley (MWS), and HP Lovecraft (HPL). The task is to predict the author given a specific excerpt. Our data scientist developed 12 classification models as shown in Table~\ref{tab:spooky_model}. Global Vectors for Word Representation (GloVe~\cite{pennington2014glove}) was used together with the input excerpts to derive the word embeddings as input for M8 to M11. She was interested in diagnosing and comparing these models. By quickly calculating the accuracy of all models, she found that M3 and M11 had the lowest log loss (0.489 and 0.479, respectively), indicating overall good  performance compared to others. Hence, she decided to mainly focus on these two models during her analysis.

\begin{table}[t]
  \small\sf
  \centering
  \vspace{2mm}
  \begin{tabular}{r|l|l|l}
    \toprule
    Model & Feature & Algorithm & Log Loss \\
    \midrule
    M0  & TF-IDF                 & Linear Regression  & 0.648 \\
    M1  & TF                     & Linear Regression  & 0.547 \\
    M2  & TF-IDF                 & Naïve Bayesian     & 0.595 \\
    M3  & TF                     & Naïve Bayesian     & 0.489* \\
    M4  & TF-IDF                 & SVM (rbf kernel)   & 0.737 \\
    M5  & TF                     & SVM (rbf kernel)   & 0.737 \\
    M6  & TF-IDF                 & XGBoost			  & 0.789 \\
    M7  & TF                     & XGBoost			  & 0.783 \\
    M8  & Word Embedding (GloVe) & XGBoost            & 0.703 \\
    M9  & Word Embedding (GloVe) & MLP                & 1.174 \\
    M10 & Word Embedding (GloVe) & LSTM               & 0.587 \\
    M11 & Word Embedding (GloVe) & Bidirectional LSTM & 0.479* \\
    \bottomrule
  \end{tabular}
  \vspace{1mm}
  \caption{Models used in the spooky author identification task.}
  \label{tab:spooky_model}
  \vspace{-4mm}
\end{table}

\subsubsection{Model Comparison and Performance Inspection}
\begin{figure}[t]
 \centering
 \includegraphics[width=\columnwidth]{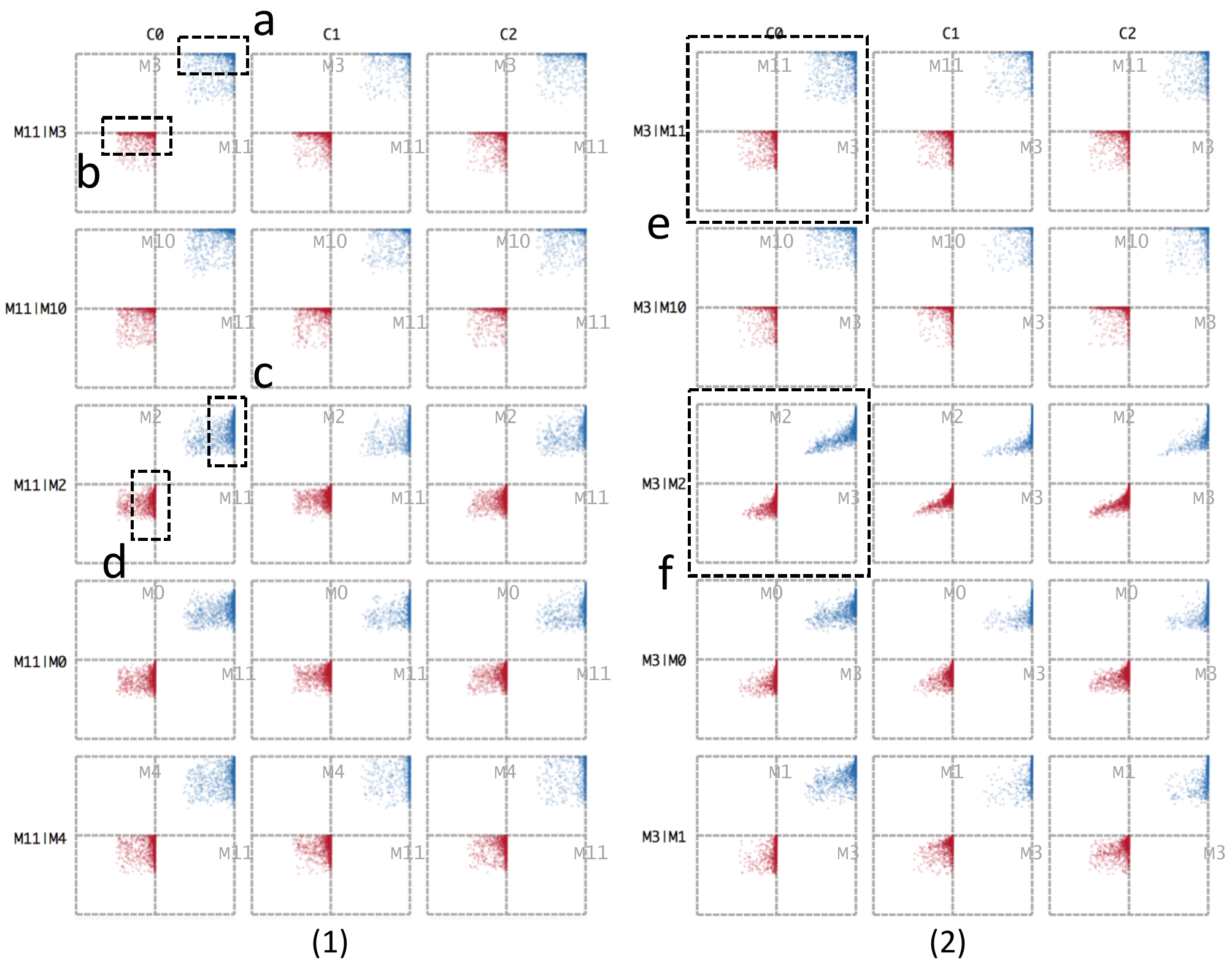}
 \caption{(1): Model performance comparison between M11 and others. The density distributions of the correctly predicted instances indicate that M3 and M10 are more confident than M11 (a, b), which is more confident than M1 and M2 (c, d). (2): Model performance comparison between M3 and other models. M3 has a higher degree of model complementarity with M10 and M11 (e). In contrast, M3 outperforms M0, M1 and M2 on a majority of instances, indicating a lower degree of complementarity (f).
 }
 \label{fig:inspection}
\end{figure}

The user loaded all models and created a mapping between columns and classes ($C_0$: EAP, $C_1$: HPL, $C_2$: MWS). She chose to use the X axis to encode M11 and the Y axis to encode other models for comparison, and filtered to show instances where both models generated correct results (true positive and true negative instances) as shown in Figure~\ref{fig:inspection}(1). From a high-level perspective, she identified that overall the instances in Q1 had a higher prediction score than those in Q3. This fact was consistent with her prior knowledge: the model predicts the class that has the highest probability among all classes. Through visual inspections on the individual cells, she identified that although both models predicted correctly, their prediction scores varied. In Figure~\ref{fig:inspection}(1), the first (M11, M3) and second row (M11, M10) show a dense horizontal line close to the top right corner of Q1 and Q3 (rectangle a and b), indicating that although the overall accuracy of M11 was higher than M3 and M10, those two models generated a higher prediction score than M11. In comparison, the third (M11, M2), fourth (M11, M0) and fifth row (M11, M4) show a dense vertical line (rectangle c and d), indicating M2, M0, and M4 were less confident than M11.

She continued to examine M3 and identified that the model pair (M3, M11) presented a distribution of a rectangular shape (rectangle e), shown in Figure~\ref{fig:inspection}(2). This reflected the fact that the number of instances on which M3 outperformed M11 and on which M11 outperformed M3 were relatively comparable, indicating the two models could potentially be combined as a model ensemble to improve the overall confidence. Since M3 was a Naive Bayes model and M11 was a deep neural network model (LSTM), this was also resonant with her domain knowledge that models of different working mechanisms have a higher degree of complementarity and are more likely to be used to create a model ensemble. In contrast, model pairs between M3 and M0, M1, M2 showed a distribution of a triangular shape (rectangle f) below the diagonal line ($y = x$), indicating that M3 outperformed M0, M1, and M2 on a majority of instances. Combining these models with M3 for model ensemble may yield little improvement.

\subsubsection{Identification and Explanation of Erroneous Instances}

After a high-level comparison of M11 and M3, the scientist was interested in which instances one model was erroneous while the other was correct. She configured the model comparison overview so that the X axis represented M11 and the Y axis represented other models (i.e., the first row represented model pair M11 and M3). Furthermore, she filtered to show instances where model X was wrong while model Y was correct. The small multiple view showed two separate sets of data: instances (blue) in the second quadrant (false negative for M11) and instances (red) in the fourth quadrant (false positive for M11). The false positive instances that had relatively high prediction score, which is defined as a \textit{symptom} set, were of particular interest to her since the model was very confident about its erroneous decision. She applied a lasso selection to the corresponding instances in the cell corresponding to the model pair (M11, M3) and class $C_1$, with the same instances highlighted in adjacent cells (blue) corresponding to $C_0$ and $C_2$. This indicated that the wrongly predicted instances belonged to both $C_0$ and $C_2$. Furthermore, the background color of the quadrant in the first column was darker than the one in the third column, suggesting that a majority of instances belonged to $C_0$. Hence, the brushing and linking interaction in \techname\ allows the user to not only have a better understanding of between-class confusion, but also examine the confidence of these decisions at a fine-grained level.

She then investigated the features contained within the selected \textit{symptom} set in Figure~\ref{fig:interpreter}(2). The KL-divergence graph clearly showed that the selected instances had a more similar distribution with class $C_0$ since the area of the line chart corresponding to $C_0$ was the largest (Figure~\ref{fig:interpreter}b). This was resonant with the fact that most of the selected instances belonged to $C_0$. She was curious about why M11 wrongly predicted the instances as $C_1$. By sorting the features based on the term frequency within the selected set, she identified that the word \textit{old} was the most frequent word. She further sorted based on class $C_1$ and noticed that \textit{old} was intensively used within that class, shown in Figure~\ref{fig:interpreter}(3). This indicated that although \textit{old} was a representative term within $C_1$, it may not be a discriminative feature among all three classes. This result recommended to her to potentially remove the feature \textit{old} or reduce its weight during the learning process. In essence, the feature interpretation view enables the users to dive into feature-wise comparison and generate potential explanations for the \textit{symptom} set of interest. These explanations can then be integrated to improve the feature set in order to validate the hypothesis and refine the model.

\subsection{Regression}
In this section, we showcase how users can leverage Manifold for regression model analysis. We use the Bike Sharing Demand dataset from Kaggle, which consists of 11000 training and 6500 test data points and features listed in Table \ref{table:regression-features}. We asked the user to build models to predict the demand, i.e., the total number of bikes rented with the Root Mean Squared Logarithmic Error (RMSLE) as the evaluation metric. Early exploratory data analysis revealed a strong correlation between features \textit{casual}, \textit{registered} and the prediction target \textit{count} (corr = 0.67, 0.98, respectively). The user hence first removed these two features to prevent data leakage. Then, she iterated with five commonly used regression models with the default hyperparameters in Scikit-learn~\cite{pedregosa2011scikit} to start with, yielding an initial set of results as depicted in Table~\ref{table:regression-models}.

\begin{table}
  \small\sf
  \centering
  \vspace{2mm}
  \begin{tabular}{r|l|l}
    \toprule
    Feature    & Type        & Description \\
    \midrule
    datetime   & datetime    & hourly date + time stamp \\
    season     & categorical & spring, summer, fall, winter \\
    weather    & categorical & clear, mist + cloudy, light snow, heavy rain \\
    holiday    & boolean     & whether the day is considered a holiday \\
    workingday & boolean     & whether the day is non weekend nor holiday \\
    temp       & numerical   & temperature in Celsius \\
    atemp      & numerical   & "feels like" temperature in Celsius \\
    humidity   & numerical   & relative humidity \\
    windspeed  & numerical   & wind speed \\
    casual     & numerical   & number of on-registered user rentals initiated \\
    registered & numerical   & number of registered user rentals initiated \\
    count      & numerical   & number of total rentals \\
    \bottomrule
  \end{tabular}
  \vspace{1mm}
  \caption{Feature table of the bike sharing demand prediction dataset.}
  \label{table:regression-features}
\end{table}

\begin{table}
  \small\sf
  \centering
  \vspace{2mm}
  \begin{tabular}{r|l|l}
    \toprule
    Model & Algorithm & RMSLE (lower is better) \\
    \midrule
    M0 & Linear Regression & 0.968 \\
    M1 & K-Nearest Neighbor & 0.731 \\
    M2 & Random Forest & 0.364 \\
    M3 & Gradient Boost Tree & 0.357 \\
    M4 & Neural Network (Multilayer Perceptron) & 1.159 \\
    \midrule
    M5 & Ensembled model + derived features & 0.341 \\
    \bottomrule
  \end{tabular}
  \vspace{0.2mm}
  \caption{Models used in the bike sharing prediction task.}
  \label{table:regression-models}
  \vspace{-2mm}
\end{table}

\begin{figure*}[t]
 \centering
 \includegraphics[width=.88\textwidth]{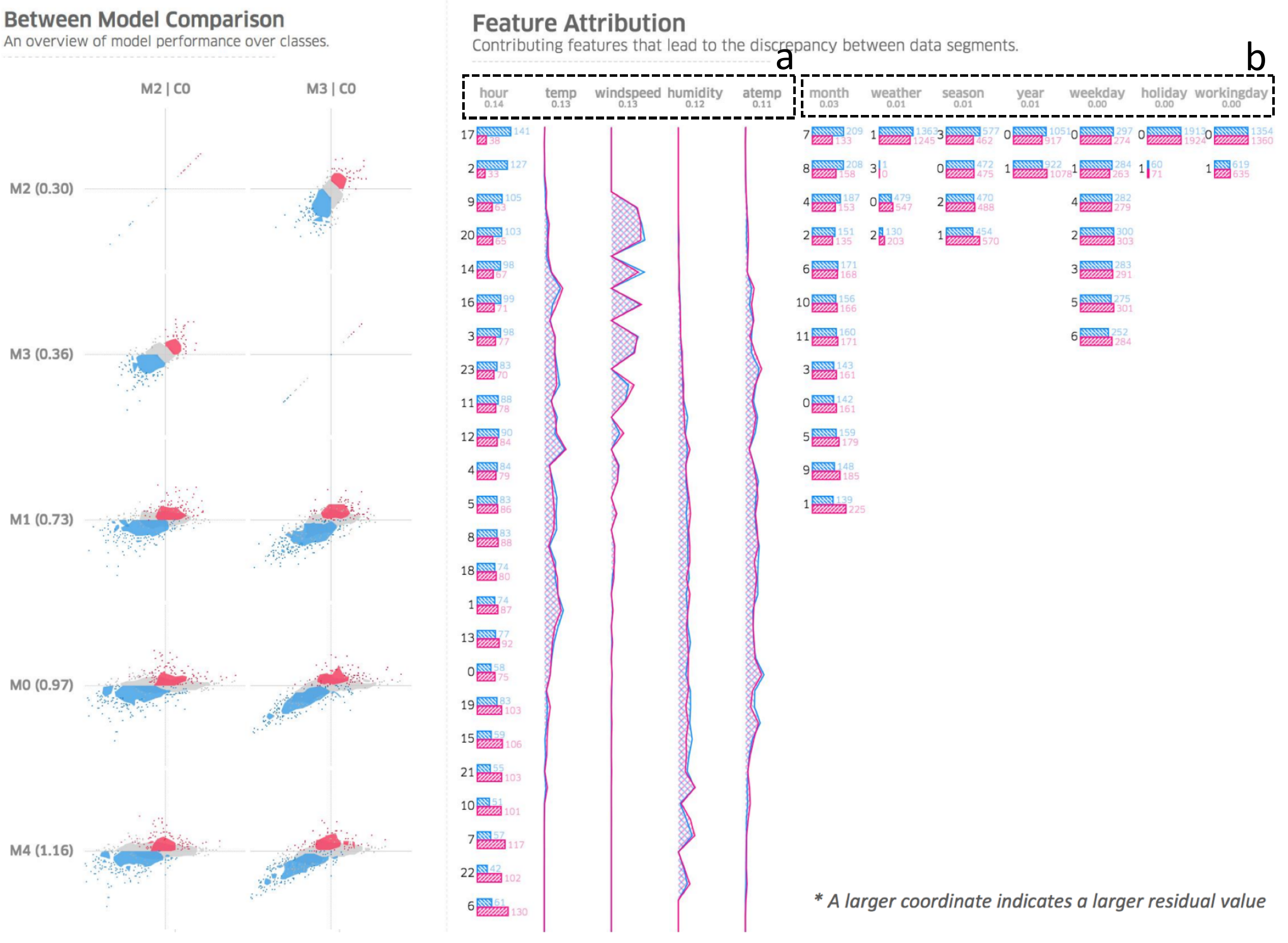}
 \caption{Left: Model performance comparison between model pairs for the bike sharing regression problem. Two instance subsets where both top two performing models (M2 \& M3) generated over-predicted (pink) and under-predicted (blue) values were selected for further study. Right: the contributing features sorted by the divergence between the value distributions of the two instance subsets.}
 \label{fig:regression-result}
\end{figure*}

\subsubsection{Model Comparison and Performance Inspection}
Looking at the RMSLE scores, the user realized model M2 and M3 performed significantly better than the other three models. She thus chose M2 and M3 as the candidate models, and encoded the data coordinates with the residual error $\epsilon = predict - actual$ in the model comparison view. As such, data instances with better predictions are projected closer to the origin of the cell, and the four quadrants correspond to $[++, -+, --, +-]$, where $+$ and $-$ depict over-prediction and under-prediction of the model, respectively.

From the comparison overview in Figure~\ref{fig:regression-result}, she confirmed both M2 and M3 had narrower distributions near the origin. She also noticed that data instances were more widely spread in the third quadrant than the other three, indicating the models tended to under-predict.

\subsubsection{Reasoning via Feature Attribution}
Based on the above observations, the user sliced out two subsets where M2 and M3 both over-predicted (red) and under-predicted (blue) in Figure~\ref{fig:regression-result}. She then derived the contributing features to the difference of the two subsets. At a glance, she identified that most categorical and binary features, such as \textit{month}, \textit{weather}, and \textit{season} (rectangle b), had low divergence scores (shown under the feature name), indicating they were not the differentiating factors of the two subsets. In contrast, features such as \textit{hour}, \textit{temp}, and \textit{windspeed} (rectangle a) had relatively large divergence, indicating they are potentially useful for deriving new features that capture the residuals.

\color{\revcolor}
\subsubsection{Feature Engineering for Model Improvement}
In order to incorporate the derived feature attribution for model iteration, the user proposed to build a stacking model that learned to reduce the residual errors. She utilized the distributions of the two subsets selected earlier, and defined feature encoders $\{F_{i}\}$ as functions corresponding to the original features $\{f_{i}\}$ with divergence larger than a user-defined threshold. $F_{i}$ maps an input value to the difference of the two distributions draw from $f_{i}$. Using the feature \textit{hour} as an example, the feature encoder generates the mapping $\{17: 103, 2: 94, ..., 22: -60, 6: -69\}$, respectively. The intuition behind such encoding was Gradient Boosting: given the user-defined data slices of interest, the user explicitly encoded their discrepancies as new features, and created a new stacking model to learn and eliminate the residuals. As a result, the RMSLE value was reduced from 0.357 to 0.341.

While Gradient Boosting is one of the various ways of incorporating such user-generated features, it demonstrates a complete human-in-the-loop workflow that is actionable for model improvement.


\subsection{Domain Expert Feedback}
\label{sec:domain}
\color{\revcolor}
\techname\ was assessed by several machine learning practitioners from a ride-sharing company, including two AI researchers focusing on text classification, three data scientists from multiple divisions (regression, risk analysis, and forecasting), and five engineers working on the machine learning platform and infrastructure. They participated in the machine learning development pipeline for different company-specific purposes and iterated and debugged their models on a daily basis.

These domain experts pointed out that the entry point of their debugging process was identifying the erroneous instances on which the model performance degraded the most. Commonly adopted performance measures such as F-measure, ROC (or AUC), and confusion matrices are usually oriented towards a single model at a coarse-grained level.
When multiple models were involved in the analysis, they often had to investigate different models independently and then combine or compare the results to form a comprehensive understanding. They admitted that this process could be tedious and inefficient even with a small number of models (i.e., 4 or 5) due to two reasons. First, the users required complex logical operations to navigate in the high-dimensional space formed by multiple models and to slice the subset of interest. Second and more importantly, these coarse-grained summaries lacked the ability to provide more fine-grained suggestions for data slicing and filtering (i.e., the slicing threshold of the prediction score). Therefore, the users had to either rely on empirical experience or manually investigate a small number of instances to gain more concrete knowledge. Hence, they all agreed that combining the results of the model pairs into the same visual display and slicing the instance space based on pair-wise correctness and confidence helped overview the model comparison and identify suspicious instances and their density distributions more efficiently. As one data scientist commented, ``this visual mapping strategy allows me to use different models as different lenses to examine data, and see how differently these lenses perform on different subsets''. An AI researcher commented: ``the visual design provides a unique and useful way to organize and separate information, and can be applied to a variety of use scenarios.''

Two data scientists noted that the Cartesian-based mapping strategy generated visual confusion when initially presented to them without training. For example, they found it difficult to understand the probability score in the negative half of the coordinate system, which was used to encode the Boolean prediction result (relevant or irrelevant) rather than a ``negative probability'' in our encoding scheme. However, once they became familiar with the coordinate encoding, they found that this design was more effective than other alternatives (Section~\ref{sec:design_alterative}) in terms of interactively selecting instances of interest (i.e., selecting all instances in four quadrants that are close to the origin). We acknowledge that these new visual encodings can potentially have a relatively long learning curve for novice users. However, we note that there is not a one-size-fits-all solution and training is required to familiarize the novice users with the framework and the visual design.

These users had positive feedback for the overall comparison-centric workflow in \techname\ that overviews the prediction results at the model level, selects suspicious instance subsets, and distills into actionable information based on the feature-level comparison. They noted that while investigating and debugging models, gaining feature-level insights were often the ultimate goal as they provided direct guidance for feature engineering and model iteration. Hence, they liked that \techname\ that provided the visual comparison mechanism at two orthogonal dimensions (the model level and the feature level) and guided the users to navigate and pinpoint their targeted issues down to the feature level. One AI researcher who focused on classifying textual data commented that ``\techname\ allowed me to identify the most influential n-grams that led to the discrepancy between data segments. More importantly, the entire process to gain the insight is explainable and reproducible''.

\color{black}

\section{Discussion}
We discuss the proposed approach in the following perspectives.

\subsection{Visual Analytics Rather Than Automatic Approaches}

\techname\ partitions instances based on model correctness and performance confidence. An alternative approach is to perform automatic clustering based on per-datum performance metrics in order to suggest users with an initial set of segments instead of having to examine the distribution in the quadrants. However, these clustering-based approaches pose new challenges. First, the clustering results are not always interpretable, especially the per-datum metrics derived from multiple models. Users had difficulty understanding why certain points were clustered together when there was no clear separation where one model performed better than the other. This also makes it difficult to act in the later ensembling stage. Moreover, preliminary evidence indicated that feature encoders generated from granular clusters tend to overfit the training set. While we still believe using semi-supervised clustering to suggest data segments is a viable approach towards less user involvement, more observations on how users slice and dice the segments need to be carefully studied, which we leave as future work.

\subsection{Contributions to the State of the Art}

As discussed earlier, many existing visual analytics solutions focus on interpreting the internal working mechanism of a specific model type~\cite{wongsuphasawat2018visualizing, ming2017understanding, kahng2018cti, liu2017towards, strobelt2018lstmvis, rauber2017visualizing}. While these approaches enable a comprehensive understanding of one model, it is usually not straightforward to transfer these insights (i.e., training strategies, hyperparameter tuning, feature/architecture engineering) to other models due to their varied working mechanisms. We design \techname\ at a higher-level scope, in the sense that we do not attempt to investigate or improve a specific model to a great extent. In contrast, \techname\ regards each model as a black box and provides comparative analysis for multiple models, allowing the end users to easily load a model into \techname\ to have an initial understanding of its performance without needing prior knowledge about the model. \techname\ not only enables model diagnosis and debugging but facilitates the understanding of a complex model as well. For example, models that generated correlated results usually have commonalities in their internal logic. Hence, end users can gain insights into the model through comparing to familiar models and utilizing prior knowledge of these well-studied models.

\subsection{Design Iterations and Alternatives}
\label{sec:design_alterative}

\color{\revcolor}
We explored several design alternatives to the scatterplot for encoding model comparison. These design iterations and feedback were mainly based on the format of interviews with our domain partners. Below we present representative design alternatives and summarize the insights gathered from the interviews.

\textbf{Whether a single and holistic visual summary or a set of individual visual slices?} From a high-level perspective, a majority of users agreed that an effective design should not attempt to encode all models into a single visual summary since this would cause severe information overload. For example, we came up with one design alternative based on confusion matrices where the multi-model results were combined within a single matrix by either displaying the value of the same model in the same position of the matrix cell or aggregating and showing summary statistics within each cell using a chart such as a box plot. The users found these alternatives to be inefficient for visual comparison as too many values were involved in the process. Instead, they preferred decomposing the holistic summary into a set of comparison units where each unit only involves a small set of models, allowing the display of more fine-grained information such as the prediction score distribution pertaining to the corresponding set. This motivated the small multiple scatterplot design in \techname.

\textbf{How many models to involve within a comparison unit?} We considered two and three models during initial design stage. A ternary plot is a natural consideration for encoding the probability distributions of the triple-class classification since the three probability values sum up to one. However, when adapted to multiple models, the predictions generated by three models are usually independent. One design we came up with was to scale the prediction scores so that they represented the relative model confidence and summed up to one. However, the major drawback is that, for example, the instances on which the three models yielded the same scores were projected to the same position in the plot no matter whether they were high or low, making the visual interpretation confusing. Another alternative extended the scatterplot design to a three-dimensional Cartesian space, which turned out to be more complicated and not scalable especially when multiple comparison units were presented to the user. Hence, we chose to not study designs involving more than two models and only focused on model pair comparison.


\textbf{Tuning the scatterplot design.} In the current scatterplot design, one difficulty mentioned earlier was to understand the probability score in the negative half of the coordinate (Section~\ref{sec:domain}). We designed an alternative that adjusted the encoding in four quadrants so that each quadrant had its origin in the lower left corner. However, several users noted that compared to the original, this design made it inefficient to select instances in a continuous range across quadrants. We integrate this design as an encoding option in \techname\ for users to choose based on their preference. To alleviate the over-plotting issue in the scatterplot, we apply a contour visualization that detects dense clusters and renders a concave hull for each cluster. For visual simplicity, we did not use multiple contour lines to visualize a fine-grained density distribution, since the data points rendered within one cell usually contains different classes (i.e., blue, red, gray). Combining fine-grained contours from multiple distributions may generate visual clutter and hinder understanding of the individual distribution. Advanced rendering and sampling methods can be integrated to remove visual confusion and preserve multi-class distributions~\cite{chen2014visual,mayorga2013splatterplots}. However, this is not the primary focus of this work, and we leave it as future work.

\color{black}
\subsection{Future Directions and Ongoing Efforts}
The scatterplot design encodes the complementarity (or diversity) of model pairs since Q1 and Q3 contain instances where the models agree and Q2 and Q4 contain instances where the models disagree. This further supports model ensemble. For example, models with varied performance on the same instance subsets can be ensembled to take advantage of both models and improve the overall performance. In contrast, models that generate correlated results on the same subsets are less useful for ensembling as the combined result remains consistent with the individual models. The model complementarity can further be quantified using specific metrics. For example, the below formula calculates a normalized score for the degree of complementarity, where larger score indicates a more substantial degree of complementarity. In this formula, $N(Qi)$ represents the number of points in Qi.
\[
    score = \frac{N(Q2) + N(Q4) - N(Q1) - N(Q3)}{N(Q1) + N(Q2) + N(Q3) + N(Q4)}
\]
Since model complementarity can be represented as a numerical value, we can further extend the small multiple view to a pixel-oriented visualization, where each cell (pixel) encodes the complementarity score of a model pair, hence accommodating model pairs and data partitions at an even larger scale.

\begin{figure}[t]
 \centering
 \includegraphics[width=\columnwidth]{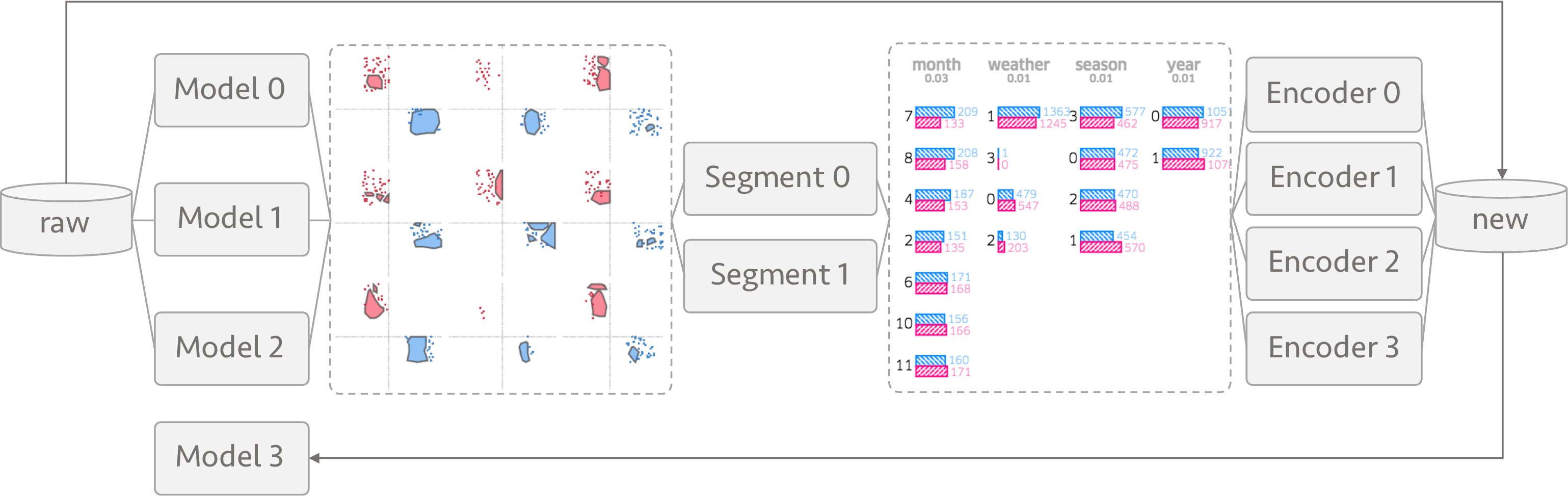}
 \caption{The workflow of \techname. Users first build a set of models, then slice out data segments of interest for feature engineering. The resulting feature encoders transform raw features into a set of new features with intrinsic structures that were not captured by the original models and help users to iterate new models and obtain better performance.}
 \label{fig:workflow}
\end{figure}

So far we have demonstrated the cyclic workflow of \techname\ as summarized in Figure~\ref{fig:workflow}. To further assess the efficacy, we have deployed Manifold as part of the core machine learning workflow within an enterprise setting. Working closely with the domain scientists who develop and tune models on a daily basis, we observe and solicit feedback on usage patterns. For instance, once familiarized with the interface, end users usually slice and compare instance subsets where models generate either consistent (e.g., the pink and blue clusters in Figure~\ref{fig:regression-result}) or converse results (e.g., the selected instances in Figure~\ref{fig:interpreter}). By collecting and generalizing these insights, domain experts can further develop semi-supervised algorithms to partition instances more intelligently and automate the reasoning process.

\section{Conclusion}

We propose \techname, a generic environment for comparing and debugging a broad range of machine learning models. \techname\ enables end users to partition instances based on model correctness and confidence, identify \textit{symptom} instances that generate erroneous results, explain potential reasons of the \textit{symptom} at the feature level, and iteratively refine the model performance.

\techname\ involves collaborations among visualization researchers and machine learning scientists that face industry-level challenges in the machine learning field. \techname\ advocates a visual exploratory approach for machine learning model development and we envision \techname\ being established as a generic platform that helps machine learning scientists manipulate complex models in a transparent and interpretable manner.

\acknowledgments{
We thank the Data Visualization Group and the AI Labs at \company\ for their support and valuable feedback.
}

\bibliographystyle{abbrv}

\bibliography{references}
\end{document}